\documentclass[twocolumn, a4paper,10pt]{article}
\usepackage[top=2.5cm, bottom=2.5cm, left=2.0cm, right=2.0cm, columnsep=0.8cm]{geometry}
\usepackage{lineno}
\usepackage{enumitem}
\usepackage[hidelinks]{hyperref}
\usepackage{boxedminipage}
\usepackage{nopageno}
\usepackage{graphicx}
\usepackage{natbib}
\usepackage[font=it]{caption}
\usepackage[usenames,dvipsnames]{xcolor}
\usepackage{listings}
\usepackage{caption}
\usepackage{subcaption}
\usepackage{soul}
\usepackage{array}
\usepackage{amsmath}
\usepackage{float}
\usepackage{booktabs}
\usepackage{adjustbox}
\usepackage{lipsum}
\usepackage{flafter}

\setlist{itemsep=-0.1cm,topsep=0.1cm,labelsep=0.3cm}
\setenumerate{leftmargin=*}
\makeatletter
\renewcommand\title[1]{\gdef\@title{\fontsize{12pt}{2pt}\bfseries{#1}}}
\makeatletter
\renewcommand\section{\@startsection{section}{1}{\z@}{3pt}{3pt}{\normalfont\large\bfseries}}
\makeatletter
\renewcommand\subsection{\@startsection{subsection}{1}{\z@}{\z@}{\z@}{\normalfont\normalsize\bfseries}}
\makeatletter
\renewcommand\subsection{\@startsection{subsection}{1}{\z@}{\z@}{0.1pt}{\normalfont\normalsize\bfseries}}

\title{%
Energy Modelling and Forecasting for an Underground Agricultural Farm\\ 																							
\vspace{4pt}
using a Higher Order Dynamic Mode Decomposition Approach} 																																
\author{																																														
Zack Xuereb Conti$^1$$^2$, Rebecca Ward$^1$, Ruchi Choudhary$^1$$^2$\\ 																										
$^1$The Alan Turing Institute, London, United Kingdom\\ 																																	
$^2$University of Cambridge, Cambridge, United Kingdom\\ 																																	
\phantom{Line 7}
\phantom{Line 8} 	 														
\phantom{Line 9}}			
\date{\vspace{-0.5cm}}	

\begin{document}
\maketitle

\section*{Abstract}	
\addtocounter{section}{1}
This paper presents an approach based on higher order dynamic mode decomposition (HODMD) to model, analyse, and forecast energy behaviour in an urban agriculture farm situated in a retrofitted London underground tunnel, where observed measurements are influenced by noisy and occasionally transient conditions. HODMD is a data-driven reduced order modelling method typically used to analyse and predict highly noisy and complex flows in fluid dynamics or any type of complex data from dynamical systems. HODMD is a recent extension of the classical dynamic mode decomposition method (DMD), customised to handle scenarios where the spectral complexity underlying the measurement data is higher than its spatial complexity,  such as is the environmental behaviour of the farm. HODMD decomposes temporal data as a linear expansion of physically-meaningful DMD-modes in a semi-automatic approach, using a time-delay embedded approach. We apply HODMD to three seasonal scenarios using real data measured by sensors located at at the cross-sectional centre of the the underground farm. Through the study we revealed three physically-interpretable mode pairs that govern the environmental behaviour at the centre of the farm, consistently across environmental scenarios. Subsequently, we demonstrate how we can reconstruct the fundamental structure of the observed time-series using only these modes, and forecast for three days ahead, as one, compact and interpretable reduced-order model. We find HODMD to serve as a robust, semi-automatic modelling alternative for predictive modelling in Digital Twins.





\section*{Key Innovations}
\begin{itemize}[noitemsep]
 	\itemsep0em
	\item Physically-interpretable environmental forecasting model for an operational underground farm
\end{itemize}

\section*{Practical Implications}

Towards data-efficient and compact controllers for Digital Twins.

\section*{Introduction}

The global initiative to reduce carbon emissions and to mitigate the impact of climate change coupled with the increased availability of wireless sensors, cloud technologies and computing power (\citet{QI20213}) are triggering the emergence of innovative data-driven strategies to better manage energy consumption in buildings. One such strategy is that of a Digital Twin (DT), which could be defined as a real-time virtual representation of a physical system or process. DTs can incorporate continuously monitored data that provides feedback for optimal management of the twinned system or process. Typically, at the core of a DT is a modelling strategy, which can either be a process model or a data-driven approach whose role is to approximate the behaviour of the system or process directly from observed data. The result is a model for diagnosing, forecasting and/or controlling the twinned system or process.  DTs in buildings could be imperative for leveraging renewable technology such as battery storage, via advanced energy model-predictive control (MPC) strategies. In this paper we focus on an operational DT for an agricultural farm situated in an underground retrofitted tunnel.  We argue that predictive models for DTs when applied for building performance,  must: i) generalise robustly across seasons with least number of tunable parameters, ii) offer model compactness, and iii) provide physical interpretability into the mechanics governing the dynamical environment.

\subsection*{Data-driven models for building performance prediction}
It is not always possible to specify equations for complex physical phenomena. Machine learning methods have become popular with the building energy community for energy consumption and performance prediction because they are relatively easy to implement and can infer complex input-output correlations between variables directly from data in an automatic manner, via black-box learning algorithms i.e. do not require knowledge or specification of Physics principles. Traditional data-driven methods include multiple linear regression (\citet{Ciulla2019}), artificial neural networks (\citet{Lu2022}), random forest (\citet{Wang2018}), and others (\citet{Seyedzadeh2018}).

However, real-world building environments experience influence from external signals (e.g. weather, occupancy, etc.,) whose behaviour may vary inconsistently over time (e.g. transient behaviour). In such cases,  traditional data-driven models tend to suffer from generalisation instability, for example when generalising for an a-typical weather scenario (\citet{ghadami2022data}). Such generalisation limitations stem from the fact that the architecture of purely data-driven models does not correspond with the mechanistic structure governing the dynamics underlying the observed time-series data, which in turn, limits diagnostic capabilities due to limited physical interpretation. These reasons may impede the application of traditional data-driven methods in real-world DTs; for example, in a divergent forecast due to model generalisation instability can lead to destabilisation of a controlled thermal environment. 

\subsection*{A (higher order) dynamic mode decomposition approach}

In this paper we argue for the need of a shift in real-world building energy modelling from traditional black-box data-fitting approaches towards methods that attempt to infer physically meaningful structure from the data where less 'blind' dependency on data volume is necessary.  More specifically, we focus our efforts on first understanding the modal characteristics of the farm's environmental quasi-periodic dynamical behaviour via mode decomposition methods and subsequently build simplified yet interpretable models for short-term forecasting. We adopt a Dynamic Mode Decomposition approach, which is a reduced order modelling technique that attempts to recover the governing dynamics directly from signal data. DMD's recent success for gaining physical insight of dynamical systems across a wide variety of applications, coupled with the ease to analyse experimental and numerical data, has led to the development of various extensions that improve the technique's robustness to cater for certain applications. One such DMD-extension is Higher Order Dynamic Mode Decomposition (HODMD), which caters for robust decomposition in noisy experimental data conditions such as sensor data \citet{Clainche2017}. HODMD is an equation-free, data-driven approach typically adopted by the fluid dynamics community to extract meaningful low-order structures that govern highly non-linear behaviour such as fluid, from data and are subsequently used to reconstruct the fluid field. 

\subsection*{Digital Twin of an operational urban-integrated farm}
This work is motivated by an operational Digital Twin of an agricultural farm called Growing Underground (GU), which exists in a retrofitted WW2 air raid shelters, 33 metres below ground level, in London (\citet{jans-singh_leeming_choudhary_girolami_2020}). The farm is equipped with a number of sensors to measure local environmental conditions at specific locations in the tunnel. Measurements from these sensors are fed directly to the DT via a data stream infrastructure and stored as time-series. In turn, the DT facilitates diagnostic tools via visualisation and analysis of the monitored data in real-time, in the form of interactive 3D visualisation, interactive charts and downloadable csv data. Subsequently, the time-series data is used to train a data-driven model to forecast temperature and relative humidity conditions for the following 24hours. The forecasts are necessary to maintain a controlled environment hence, secure a constant crop yield. 

The current DT adopts a seasonal auto-regressive integrated moving average (SARIMA) model, as discussed in previous work (\citet{RWard2023}). In this paper we illustrate how a HODMD-based approach could contribute to modelling and short-term forecasting environmental behaviour in the underground farm, as a data-efficient, physically interpretable and more compact alternative to the current implemented SARIMA model, for diagnostic and future farm-control purposes. More specifically, we focus on modelling and forecasting vapour pressure deficit (VPD), which is a diagnostic measure widely adopted in agriculture to monitor crop growth.  

\subsection*{Document structure}
The paper is structured as follows: in the next section we provide details about the sensing infrastructure installed in the farm, followed by the HODMD methodology and how we apply it for analysing and forecasting monitored VPD signals. Lastly, we apply the HODMD approach to historical monitored data from three seasonal scenarios in a retrospective study, to show how we can infer consistent modal structures and use these to forecast ahead in time.

\begin{figure*}[htp]
\centering
\includegraphics[width=0.4\textwidth,]{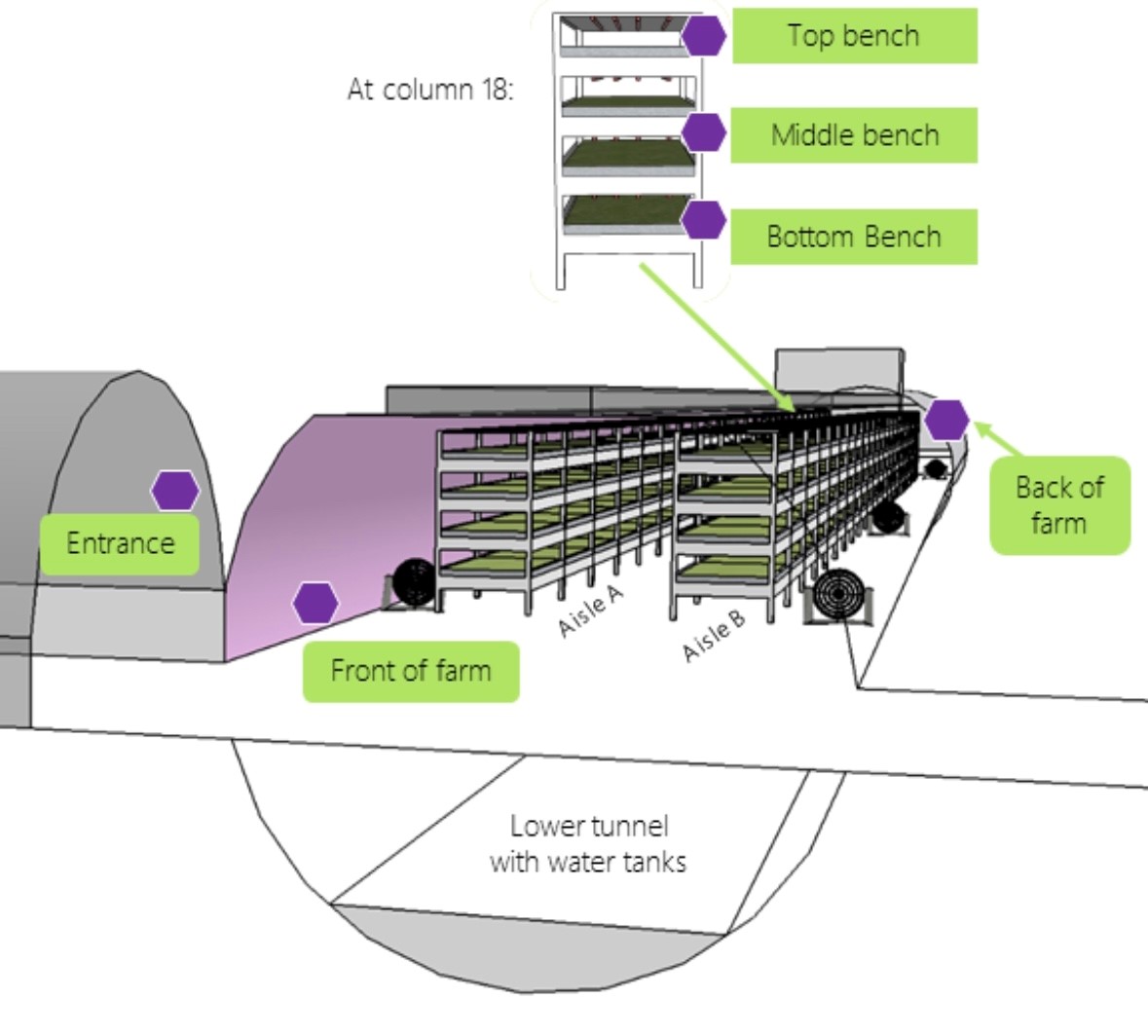}\hfill
\includegraphics[width=0.5\textwidth]{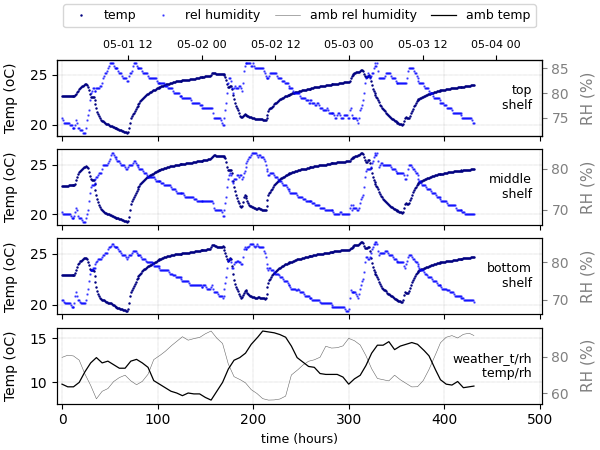}\hfill
\caption{Cross-sectional figure of the underground farm indicating the sensor locations for this study (left).  Temperature and relative humidity observed at the centre of the farm (right). }
\label{fig:fig01}
\vspace{-16pt}
\end{figure*}

\section*{Monitoring an operational underground farm}

In summary,  GU is an artificially-lit hydroponic farm producing micro-greens composed mainly of pea-shoots, basil, coriander, parsley, salad rocket,  pink radish and mustard plants. The site consists of two parallel tunnels, running on two levels and spanning approximately 400m. The growing area of the farm consists of two aisles (A and B), with each aisle consisting of 4 tray stacks, respectively, totalling to 528m2 of farming area (Figure~\ref{fig:fig01}, left). The farm tends to be occupied in the mornings (6-8am) to harvest fresh produce, an in the evenings (3-4pm),  to exchange the crops. The artificial lights are switched on and off daily to mimic a diurnal sunlight cycle with 16 hours of daylight and 8 hours of night, reflected by the periodicity in the measured temperature and relative humidity time-series (Figure~\ref{fig:fig01}, right). The ultimate goal of the farm is to control environmental conditions to maintain steady crop growth rates that satisfy (promised) harvest targets, with as little as crop `damage' as possible. The latter is heavily influenced by the environmental conditions of the farm for which reason sensors monitor temperature and relative humidity at several locations throughout the farm. The monitored values are used to calculate vapour pressure deficit (VPD).


VPD is one of several methods crop growers use to measure the humidity in a greenhouse or growing environment. It allows growers to gauge the impact of humidity on plant growth and development. VPD can be described intuitively as the `remaining capacity' for more humidity in the air at a given temperature. The VPD can be calculated by subtracting the water vapour pressure in the air from the saturated vapour pressure (SVP) at a specific temperature, as follows: 
\begin{equation}\label{eq:1}
  SVP = 610.78 * e^{T / (T +237.3) * 17.2694}
\end{equation}
\begin{equation}\label{eq:2}
  VPD = SVP * (1-RH/100)
\end{equation}

\section*{VPD modelling and forecasting via HODMD}
HODMD is a reduced order model (ROM) technique recently introduced by \citet{Clainche2017} as an extension of an emerging ROM technique in the fluid dynamics community called dynamic mode decomposition (DMD). In this paper we apply it to VPD as a function of observed temperature and relative humidity sensor data. 

\subsection*{Introduction to dynamic mode decomposition}

DMD is an `equation-free' technique developed by \citet{Schmid2008} which aims to infer physically-meaningful spatio-temporal structures of the dynamics from measured data. This is a significant advantage over typical data-driven modelling methods which operate as black box models. More specifically, DMD retrieves spatially-correlated structures with similar behaviour in time. These structures, also referred to as DMD modes, correspond strongly to the eigenvectors of the state space governing the underlying dynamic system. DMD is very similar to other decomposition methods such as principal orthogonal decomposition (POD), main difference being that DMD modes are based on their dynamics rather than the energy content. A POD mode may contain a continuous frequency spectrum, while each DMD mode is characterized by a single frequency thus, rendering DMD an efficient way to identify dominant frequencies (\citet{Wu2019}). While DMD facilitates a reduction in the order of dimensions based on a reduced set of modes, it also provides a model for the interaction of those modes in time. 

DMD decomposes the spatio-temporal data \(v_{k},\)  as an expansion of Fourier-like modes: 
\begin{equation} \label{eq:3}
v_{k} = \sum_{m=1}^{M}\alpha_{m} u_{m}  e^{(\delta_{m}+i\omega_{m})t_k}
\end{equation}
where \(u_{m}\)  are the DMD modes. These modes are weighted by an amplitude \(a_{n}\), they grow or decay in time by a growth rate \(\delta_{n}\), and oscillate with a frequency \(\omega_{n}\). The number of DMD modes $M$ required to characterise (and reconstruct) the signal, can be referred to as the spectral complexity, while K is the temporal dimension, which generally represents the number of snapshots available for the decomposition. 

In order to obtain the expansion in (\ref{eq:3}), the observed data is first organised as a set of so called snapshots, which are observed instances at equispaced points in time \( \Delta t \) and stored as a matrix of row vectors $X_{k}$, whose columns $(x_{1},x_{2},... ,x_{m-1})$ represent snapshots of the measured system, as shown in (\ref{eq:4}). In our farm case study, the columns of the snapshot matrix represent sensor measurements at different spatial locations at a single time step. A consecutive snapshot matrix $X_{k+1}$ is formulated by offsetting $X_{k}$ by one time step \( \Delta t \) as shown in (\ref{eq:7}), this is referred to as the time-delayed matrix. 

\newcommand*{\vertbar}{\rule[-1ex]{0.5pt}{2.5ex}}
\newcommand*{\horzbar}{\rule[.5ex]{2.5ex}{0.5pt}}

\begin{equation}\label{eq:4}
X^{K-1}_{1} = 
\left[
  \begin{array}{cccc}
    \vertbar & \vertbar &        & \vertbar \\
    x_{1}    & x_{2}    & \ldots & x_{K-1}    \\
    \vertbar & \vertbar &        & \vertbar 
  \end{array}
\right]
\end{equation}

\begin{equation}\label{eq:5}
X^{K}_{2} = 
\left[
  \begin{array}{cccc}
    \vertbar & \vertbar &        & \vertbar \\
    x_{2}    & x_{3}    & \ldots & x_{K}    \\
    \vertbar & \vertbar &        & \vertbar 
  \end{array}
\right]
\end{equation}

The main principle behind the original DMD algorithm, is to learn a linear matrix operator $\hat{A}$ that updates $X_{k}$ to $X_{k+1}$, as specified in (\ref{eq:6}). The operator $\hat{A}$ contains the dynamical information about the system. It is said that the recovered $\hat{A}$ by DMD is a finite-dimensional approximation of the so-called Koopman linear matrix operator (\citet{kutz2016dynamic}),  which is a `theoretically' infinite-dimensional operator to approximate non-linear behaviour. 

\begin{equation}\label{eq:6}
	X^{K}_{2}=\hat{A}X^{K-1}_{1}
\end{equation}

The dynamics of real-world systems such as thermal and relative humidity exchanges in the underground farm, experience forcing effects due to foreign influences such as the outdoor climate, occupants' behaviour, and other sources. These influences result in what we observe as noisy measurement signals, and pose a challenge for learning data-driven models that can temporally generalise robustly. In more detail, real-world physical systems such as the retrofitted farm, are often governed by a broad spectrum of modes, which can be difficult to capture by traditional black-box methods if the data is not representative enough. Even DMD tends to fail in capturing the dynamics for such cases, particularly when the spectral complexity of observed signals is of an order larger than the spatial complexity. In our context, the spatial complexity is directly influenced by the number of sensors available. In other words, the sparsity of the sensors is not sufficient to reveal the broad range of modes governing the observed dynamics. For these reasons, we adopt HODMD, developed by \citet{Clainche2017} as an extension to DMD to handle such scenarios by exposing a broader spectrum of modes than standard DMD.

\subsection*{Higher order dynamic mode decomposition}

\citet{Clainche2017} argue that HODMD is very useful to analyse periodic and quasi-periodic phenomena in pattern forming systems, which is fitting for analysing the farm's environmental behaviour given its periodic operation. HODMD can be interpreted as the regular DMD algorithm applied to a time-delayed sliding-window approach. Based on Taken's delayed-embedded theorem, HODMD relates $d$ time-delayed snapshots using higher-order Koopman assumption defined as: 
\begin{align}\label{eq:7}
	& {X}_{{k}+{d}}= A_{1}x_{k}+A_{2}x_{k+1}+...+A_{d}x_{k+d-1} \\
	& \mbox{for } k=1, ..., K-d \nonumber
\end{align}

where $d\geq 1$ is a tunable parameter. Suitable values for $d$ depend on the dynamics, the sampled time interval and the sampling frequency. $d$ scales with the number of snapshots such that if $K$ is doubled, so is $d$ (\citet{Clainche2017}). The higher order expansion in (\ref{eq:7}) can be obtained algorithmically in two main steps: first a dimension reduction via SVD followed by the DMD-d algorithm, as follows. 

In the first step, singular value decomposition (SVD) is applied to the snapshot matrix $X_{k}$ resulting in the following decomposition of the spatio-temporal data:

\begin{equation}\label{eq:8}
	\mbox{SVD: } X_{k} \approx {U} \Sigma {V}^{T}  
\end{equation}
where, 
\begin{equation}\label{eq:9}
	\hat{X}_{k}= \Sigma {V}^{T}
\end{equation}

and where, $\Sigma$ is a diagonal matrix containing the retained SVD singular values sorted in decreasing order, $U$ is a matrix containing the orthonormal SVD spatial modes, $V$ contains temporal SVD modes, and $\hat{X}_{k}$ is the truncated snapshot matrix. The truncated SVD facilitates the reduction of the original snapshot data into a series of linearly independent vectors of dimension $N$ (where $N$$<$$J$ is the spatial complexity). The number of $N$ retained modes in the expansion is chosen by means of a user-tunable tolerance $\varepsilon_{1}$.
 

In the next step, we apply the DMD-d algorithm directly on the reduced snapshot matrix $\hat{X}_{k}$, which relates $d$ subsequent snapshots as expressed in (\ref{eq:7})  in the following way:
\begin{equation}\label{eq:10}
	\hat{X}_{d+1}= \hat{A}_{1}\hat{X}^{K-d}_{1}+\hat{A}_{2}\hat{X}^{K-d+1}_{2} +...+\hat{A}_{d}\hat{X}^{K-1}_{d}
\end{equation}

where, $\hat{A}_{l}$ (with $1 \leq l \geq d$) are the reduced Koopman operators, obtained modifying the standard Koopman operators via $\hat{R}_{l}=U^{T}R_{l}U$. It is noteworthy to point out that when $d$=1 (DMD-1), the high-order expression in (\ref{eq:10}) is equivalent to the standard DMD algorithm. Subsequently, the multiple Koopman operators $\hat{A}_{1}, ... ,\hat{A}_{K}$ are then combined into a single matrix, with which the eigenvalue problem can be solved to obtain the DMD modes \(u_{m}\), their corresponding amplitudes \(a_{n}\), frequencies \(\omega_{n}\), and growth rates \(\delta_{n}\). In the typical implementation of HODMD, retained DMD modes are selected based on their amplitude using a a user-specified tolerance $\varepsilon$ however, in this paper we adopt an alternative mode-selection criterion approach discussed below. Once these are obtained, it is possible to reconstruct the original time-series using the general DMD expansion in (\ref{eq:3}). For good approximations of the expansion in (\ref{eq:3}), not only is it possible to reconstruct the original data but also to predict temperature and relative humidity patterns at advanced time stages ($t_{r} \gg t_{k}$). In our case, we focus on predicting VPD.  Our HODMD implementation in this paper is based on PyDMD,  a Python package developed by \citet{demo18pydmd}.

Additionally, we adopt a recently developed algorithmic add-on to correct the effect of sensor noise in the HODMD, where the low-rank operator $\hat{A}$ is calculated using a `forward-backward' approach developed by \citet{Dawson_2016}. The latter acts as a correction filter to overcome bias from noise present in sensor data, which standard DMD tends to be sensitive to. In our results we find this to be very effective on improving the accuracy of the captured dynamics, i.e. improved time-series reconstruction. 

\subsection*{Dominant mode selection}

The expansion is further reduced by sorting the DMD modes by their dominance. In modal analysis, dominance of modes is often ranked by the corresponding amplitude of their time evolution however, we adopt an improved DMD-mode selection criterion based on the `integral contribution', as suggested by \citet{KOU2017109} where the initial condition and the temporal evolution of each DMD mode are considered for evaluating the contribution of each mode, and where numerically transient modes (very large growth/decay rates) are excluded from selection. Subsequently, the full set of ranked modes is truncated at an elbow point, which we manually determine by elimination, while observing the sensitivity to capture global features of the VPD time-series in the reconstruction and forecast across sensors and seasons.

\section*{Results}

In this section we illustrate results from the application of HODMD to analyse and forecast VPD in the underground farm using observed temperature and humidity time-series data measured via three respective sensors located at the centre of the farm. The three sensors are installed at the top, middle and bottom bench of aisle B, as illustrated by the cross-sectional drawing in Figure \ref{fig:fig01}. 

For this study we consider three scenarios: a `typical' farm scenario in winter (scenario A) and a `typical' farm scenario in summer (scenario B) where the bench stack lighting schedule, outdoor climate and farm operation were under typical conditions, and an `atypical' scenario in summer (scenario C) where the farm experienced abnormal outdoor climatic conditions. Our aim is two fold: a) to identify consistent modal structures governing the farm's VPD behaviour in the farm across both typical and atypical environmental scenarios, and subsequently b) to construct a simplified and interpretable model for short-term forecasting of VPD time-series across all three sensors. 


The experimental data is in the form of VPD time-series data having a 10-minute time-step and was computed as a function of temperature and relative humidity measurements recorded every 10 minutes, using \ref{eq:1} and \ref{eq:2}. More specifically, the training data is formed as following, scenario A: 25-01-2022 to 25-02-2022 (4459 snapshots), scenario B: 01-05-2022 to 31-05-2022 (4315 snapshots) and scenario C: 01-07-2022 to 31-07-2022 (4315 snapshots). The data was preprocessed via linear interpolation to account for any missing data due to sensor malfunction or sensor maintenance. Furthermore, the data was standardised by centring at the zero mean. We observed the latter to improve significantly the reconstruction and forecasts of the VPD.  

\subsection*{HODMD calibration}
For each scenario, the VPD time-series data was first collected into a snapshot matrix as specified in (\ref{eq:4}), where each column represents a snapshot in time of the VPD at the three sensor locations, at timestep $k$. To calibrate the tunable parameters $d$ and $\varepsilon_{1}$, HODMD was applied using three orders of tolerances $\varepsilon_{1}=0$, $\varepsilon_{1}=1e-1$ and $\varepsilon_{1}=1e-2$ across the three scenarios (Table \ref{tab:tab01}). \citet{doi:10.2514/6.2019-1531} recommend to set $\varepsilon_{1}$ at least equal or even beyond the level of noise, in their application for predicting aeroelastic flutter using real flight test data. Subsequently, we applied the method using various values of $d$ in the range $400 \leq d \geq 2000$ while recording the RRMSE of the corresponding three-day forecast, for each $d$, as tabulated in Table \ref{tab:tab01}. 

Using Table \ref{tab:tab01}, we manually identified $d=710$, $d=710$ and $d=910$ for the Winter (typical behaviour), Summer (typical behaviour) and Summer (extreme behaviour), respectively. While we note that in a DT setting, it would be ideal to limit parameter sensitivity by identifying a mutual elbow value for $d$ across all scenarios, in this study our aim is to identify coherent structures across all scenarios. Selecting $d$ in this manner is typical for HODMD applications (\citet{Clainche2017}, \citet{KOU2017109}, \citet{doi:10.2514/6.2019-1531}). Note that the overall order of magnitude of the resulting reconstruction error values range within $10^1 \leq  RRMSE  \geq 10^2$ and is considered acceptable, especially when considering the noisy environmental conditions in the farm. 


\begin{table*}[t]
\vspace{-5pt}   
\caption{Calibration study for d and for $\varepsilon_{1}$ given d=710(jan), d=710(may), d=910 (july).}
\label{tab:tab01}
\centering
\begin{adjustbox}{width=1\textwidth}
\small
\begin{tabular}{| c | c | c | c | c | c | c | c | c | c |}
\toprule
$d$  &  \multicolumn{3}{c}{January} & \multicolumn{3}{c}{May} & \multicolumn{3}{c}{July}\\
\midrule
  { } & $RRMSE_{18}$ & $RRMSE_{27}$ & $RRMSE_{23}$ & $RRMSE_{18}$ & $RRMSE_{27}$ & $RRMSE_{23}$ & $RRMSE_{18}$ & $RRMSE_{27}$ & $RRMSE_{23}$ \\
  400 &17.05 & 14.24 & 14.15 &13.16 &12.33 & 13.95 & 28.45 & 21.48 & 24.24 \\
  500 & 14.93 & 11.62 & 11.78 & 12.05 & 10.47 & 11.93 & 31.65 & 19.00 & 22.80 \\
  600 & 13.76 & 11.68 & 12.20 & 14.61 & 13.51 & 14.81 & 26.20 & 20.50 & 22.80 \\
  700 & 14.94 & 13.22 & 13.40 & 10.35 & 8.67 & 10.31 & 28.60 & 19.96 & 22.62 \\
  800 & 16.27 & 13.65 & 13.57 & 11.60 & 9.80 & 11.10 & 23.96 & 21.45 & 24.31 \\
  900 & 12.99 & 9.83 & 10.11 & 10.65 & 9.50 & 10.78 & 31.26 & 20.42 & 21.74 \\
  1000 & 13.10 & 9.75 & 10.02 & 10.56 & 8.79 & 9.95 & 19.47 & 18.43 & 22.47 \\
  1100 & 11.51 &  9.96 & 10.64 & 9.56 & 8.61 & 9.51 & 16.29 & 18.11 & 22.36 \\
  1200 & 16.04 & 13.39 & 13.34 & 8.38 & 7.29 & 8.11 &19.57 & 18.57 & 23.40 \\
  1300 & 10.95 & 7.26 & 7.73 & 8.97 & 7.77 & 8.22 & 15.26 & 17.55 & 21.94 \\
  1400 & 12.27 & 9.28 & 9.71 & 9.26 & 7.78 & 8.41 & 16.71 & 18.65 & 22.42 \\
\midrule
$\varepsilon_{1}$  &  \multicolumn{3}{c}{} & \multicolumn{3}{c}{} & \multicolumn{3}{c}{}\\
\bottomrule
  0 & 103.9 & 107 & 107.6 &12.4 &11.22 & 12.04 & 48.57 & 48.32 & 49.49 \\
  0.1 & 23.24& 15.12 & 14.77 & 12.36 & 11.22 & 12.04 & 46.81 & 22.70 & 23.21  \\     
  0.01 & 23.24& 15.12 & 14.77  & 12.36 & 11.22 & 12.04 &46.81 & 22.71 & 23.21 \\  
\bottomrule
\end{tabular}
\end{adjustbox}
\vspace{-5pt}   
\end{table*}

\subsection*{Modal analysis of VPD in the underground farm}
\begin{figure*}[htp]

\centering
\includegraphics[width=.320\textwidth]{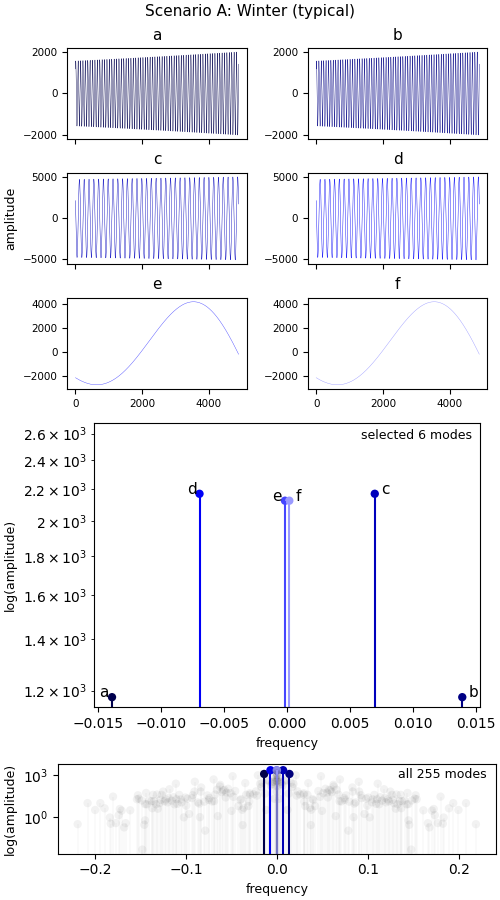}\hfill
\includegraphics[width=.320\textwidth]{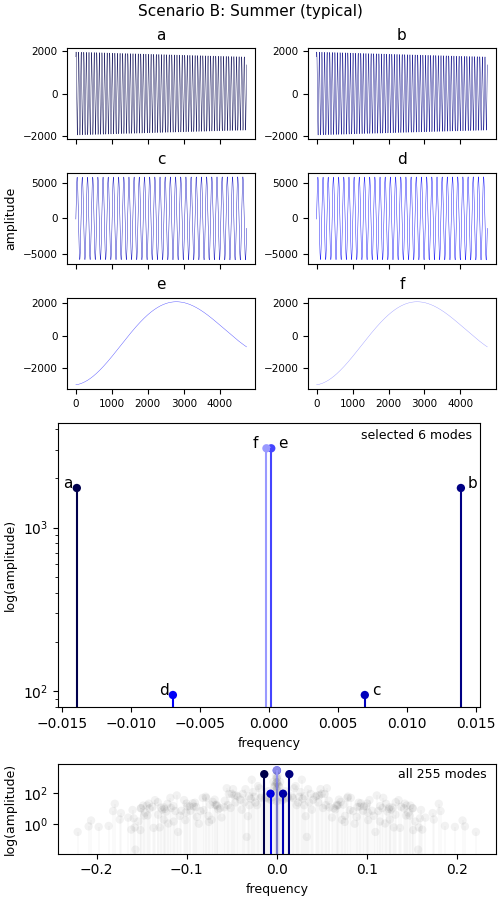}\hfill
\includegraphics[width=.320\textwidth]{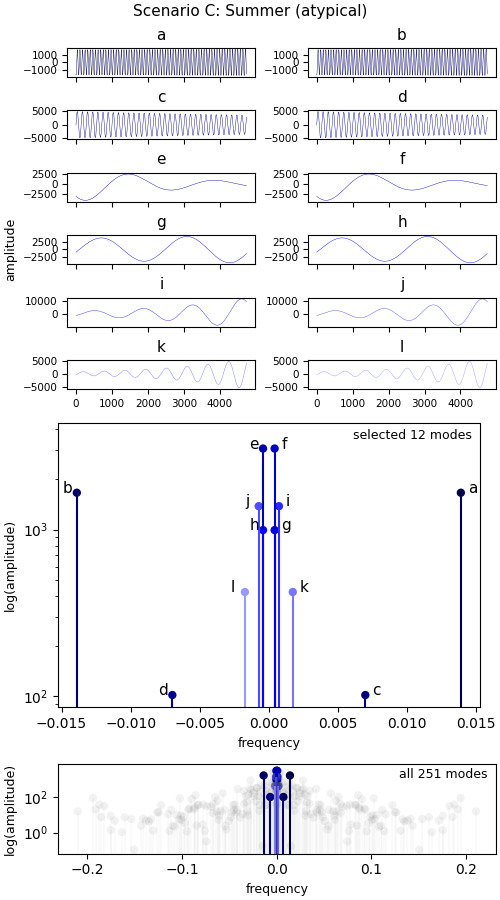}

\caption{Dominant modes and their corresponding amplitudes and frequencies (top and middle),  all modes (bottom) for scenarios DMD-710 (scenario A), DMD-710 (scenario B), DMD-910 (scenario C).}
\label{fig:figure2}
\vspace{-16pt}

\end{figure*}

We computed HODMD using the selected $d$ values corresponding to each scenario via DMD-710, DMD-710 and DMD-910, respectively. Utilising the mode-selection criterion and manual truncation discussed above, we ranked and identified the dominant modes that describe the fundamental VPD behaviour in the farm as a function of temperature and relative humidity signals, for each scenario. Figure~\ref{fig:figure2} illustrates the time-evolution of the selected dominant modes (top) and their corresponding amplitude versus frequency plot (below). In more detail, in case scenario A we identified 6 dominant modes (3 conjugate pairs) from a total of 255 modes, in case scenario B 6 dominant modes from 255 and lastly, case scenario C 12 dominant modes from 255. Selecting additional modes resulted in a 'tighter' fit in the reconstruction and forecast but for brevity's sake in this paper we favour model simplification and robustness over fit.  

\begin{figure*}[htp]

\centering
\includegraphics[width=\textwidth]{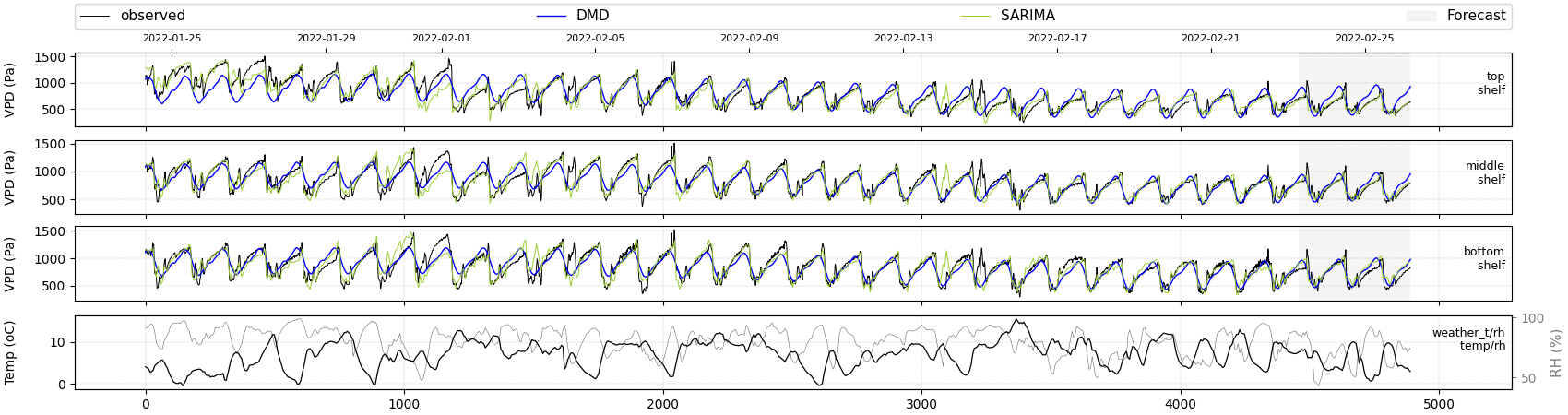}\hfill
\includegraphics[width=\textwidth]{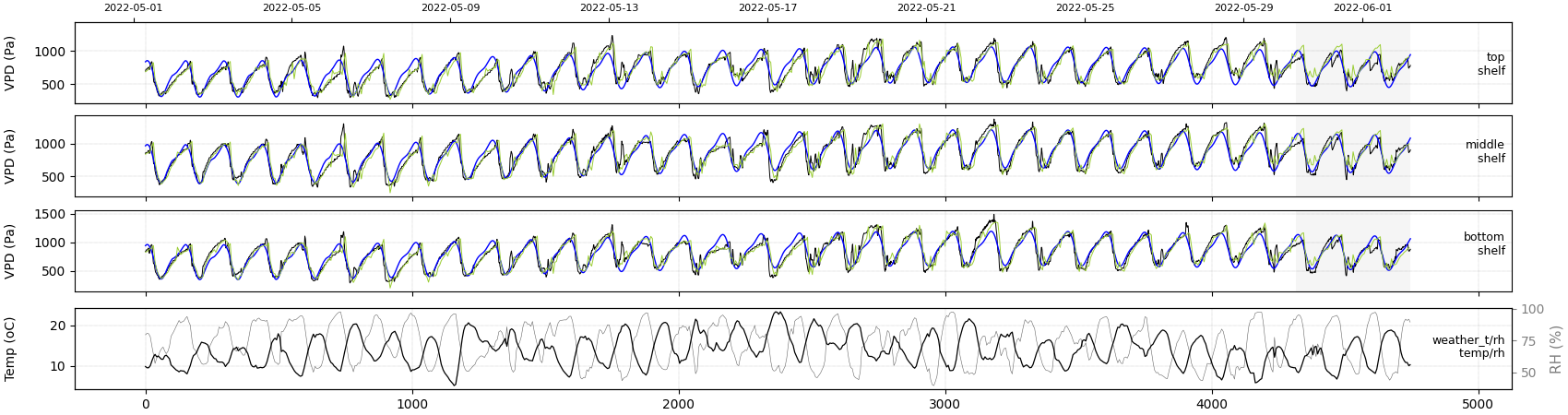}\hfill
\includegraphics[width=\textwidth]{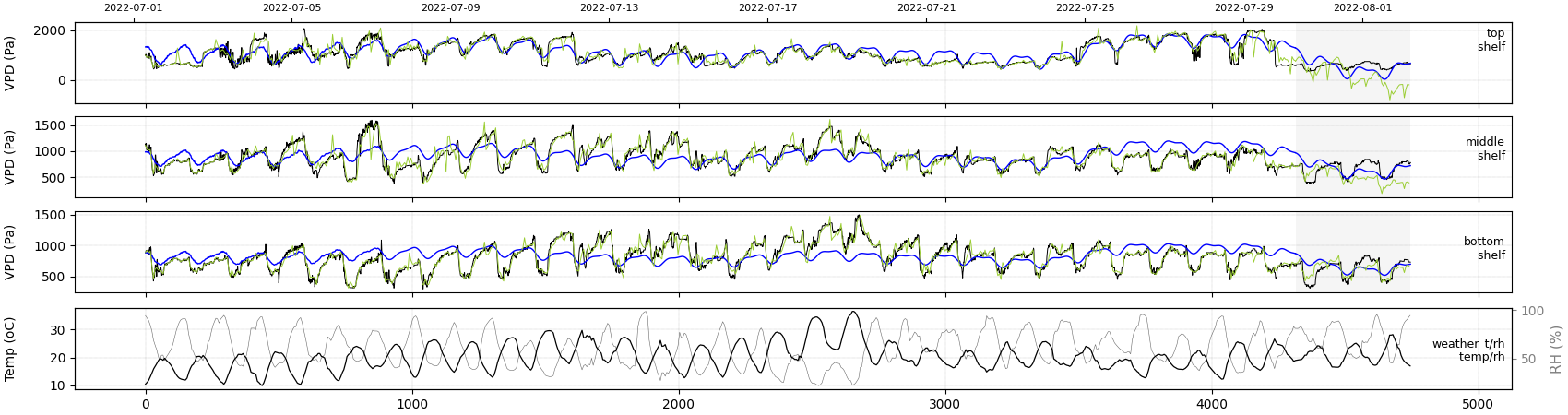}

\caption{VPD time-series reconstruction and three-day forecast for: Winter scenario A (top), Summer scenario B (middle), Summer scenario C(bottom)} 
\label{fig:figure3}
\vspace{-16pt}

\end{figure*}

Using accumulated knowledge about the farm's behaviour we were able to interpret `general' dominant mode pairs across both scenarios A and B, where environmental and operational conditions were typical. More specifically,  the frequency of mode pair $e$$f$ at +/- 0.000172Hz corresponds with the mean trend of the VPD time-series across the three sensors thus implying correspondence with the influence of external ambient temperature and relative humidity (weather). In fluid dynamics such a mode would almost correspond with what is known as the `time-average flow'. Furthermore, $c$$d$ at +/- 0.00695 Hz correspond with the periodicity of the LED lighting schedule. Lastly, we identify mode pair $a$$b$ at +/- 0.0139 Hz as miscellaneous`farm dynamics', such that it accounts for remaining general farm behaviour by elimination. We note that across both scenarios, their frequencies remain the same while their amplitudes vary, as expected due to varying forcing conditions. 

As for scenario C,  we note that further modes were required to describe the transient-like dynamics governing the VPD due to atypical outdoor climatic behaviour (a heatwave). However, we observe that the 3 fundamental mode pairs identified in A and B, also appear consistently. From  Figure~\ref{fig:figure2} it is evident that the `additional' modes  beyond these 3 pairs are concentrated around the mean frequency indicating likely correspondence with the ambient weather-related influence. We can also note from scenarios B and C in Figure~\ref{fig:figure2}, how not all dominant modes necessarily have the largest amplitudes.



\subsection*{Short-term forecasting VPD}
Using only the selected modes for each scenario, we were able to reconstruct the VPD time-series at the three vertically stacked sensor locations (top shelf, middle shelf, bottom shelf), for the three scenarios. Figure~\ref{fig:figure2}, illustrates how HODMD is able to capture and reconstruct quite well the periodic dynamics of VPD in the farm for the first two scenarios under typical behaviour, and subsequently forecasts successfully for the next three days, with a corresponding average RMSE of 0.177 and 0.118, respectively, across the three sensors. As for the third scenario, we can observe how the reconstructed signal is not a tight fit as scenarios A and B due to the transient phenomenon influencing the stable system, yet captures the quasi-periodic dynamics reflected by modes $c$ and $d$. 


\subsection*{Discussion: model robustness vs. fit}
We argue that in the context of a DT, the quality of a model is not solely dictated by the fit but rather, the interpretability and robustness of the same model to generalise across seasons and types of behaviour (e.g. stable, non-transient, etc.); one whose structure corresponds with the fundamental characteristics governing global behaviour. Therefore, while scenario C does not appear to be successful by traditional fitting standards, we find value in the stability of the forecast, even if coarse, which in a control scenario is more reliable than an unstable divergent forecast as witnessed by the SARIMA model in Figure~\ref{fig:figure3}. 

More specifically,  HODMD acts as a noise filter by decoupling arbitrary noise from the governing dynamics. This is reflected by a smoother fit at lower truncation-orders which is where the fundamental modes live. While traditional data-driven models might fit locally to the noisy signal, they are biased models. Lastly, we observed that when the system's eigenvalues tend towards stability such as in scenarios A and B, HODMD is able to forecast robustly and accurately for longer periods of time, up to 1 week, using only the three mode pairs. This is a reflection of global dynamics being captured. 

Overall, we observed that the number of modes to include for signal reconstruction and forecasting is informed by a trade-off between order-simplicity and fitting smoothness. Inadvertently, including many redundant modes can lead to over fitting and or unstable forecasts. We determined the mode truncation order for each scenario by identifying the elbow point after which global characteristics such as periodicity and dampened peaks, etc., cease to appear. For example, eliminating the  modes $e$ and $f$ in scenario A results in overall smoothing of the reconstructed signal, indicating a sensitivity in describing global behaviour. 



\section*{Conclusion}

{\setlength{\parindent}{0pt}
\setlength{\parskip}{2pt}

This paper introduced HODMD as a robust data-driven approach for modal analysis and short-term forecasting in the context of an existing Digital Twin of an underground agricultural farm. Unlike traditional data-driven methods typically adopted for building energy modelling and forecasting, HODMD attempts to infer the structure of the dynamics governing the observed signal data which in turn facilitates identification and interpretability of the dominant modes. We applied HODMD to data measured in the farm, for three seasonal scenarios. Through the study we revealed three physically-interpretable governing mode pairs consistently across all three seasonal scenarios thus, highlighting their significance for explaining and modelling the environmental behaviour in the farm. Using only these modes we could forecast robustly for three days ahead in the typical environmental scenarios while for the atypical scenario, additional modes were required due to transient behaviour dominating the signal. Furthermore, we demonstrated how HODMD demands only one compact model to forecast for multiple sensors because it is both space and time aware; a significant benefit over traditional data-driven models for MPC purposes in DTs. While the paper focused on an urban agriculture application, the HODMD-based forecasting approach can be applied to spatially model any building thermal environment where internal and external conditions are monitored. The latter may be limited by the sparsity of the sensed locations. 


\section*{Acknowledgment}

We woud like to thank the management and staff of Zero Carbon Farms Ltd and the Growing Underground Farm in Clapham, London, for their provision of unlimited access to data and information.


\bibliographystyle{BS2021}
\typeout{}
\bibliography{references_edited}
\newpage
\onecolumn

\end{document}